\documentclass[10pt,twocolumn,letterpaper]{article}

\usepackage{cvpr}              

\definecolor{cvprblue}{rgb}{0.21,0.49,0.74}
\usepackage[pagebackref,breaklinks,colorlinks,allcolors=cvprblue]{hyperref}

\def\paperID{*****} 
\def\confName{CVPR}
\def\confYear{2025}

\title{Clip4Retrofit: Enabling Real-Time Image Labeling on Edge Devices via Cross-Architecture CLIP Distillation}

\author{
Li Zhong, Ahmed Ghazal, Jun-Jun Wan, Frederik Zilly, \\
Patrick Mackens, Joachim E. Vollrath, Bogdan Sorin Coseriu \\
Robert Bosch GmbH \\
{\tt\small \{li.zhong, ahmed.ghazal, jun-jun.wan\}@bosch.com} \\
{\tt\small \{frederik.zilly, patrick.mackens, joachim.vollrath\}@de.bosch.com} \\
{\tt\small Bogdan-Sorin.Coseriu@ro.bosch.com}
}

\begin{document}
\maketitle
\begin{abstract}
Foundation models like CLIP (Contrastive Language–Image Pretraining) have revolutionized vision-language tasks by enabling zero-shot and few-shot learning through cross-modal alignment. However, their computational complexity and large memory footprint make them unsuitable for deployment on resource-constrained edge devices, such as in-car cameras used for image collection and real-time processing. To address this challenge, we propose Clip4Retrofit, an efficient model distillation framework that enables real-time image labeling on edge devices. The framework is deployed on the Retrofit camera, a cost-effective edge device retrofitted into thousands of vehicles, despite strict limitations on compute performance and memory. Our approach distills the knowledge of the CLIP model into a lightweight student model, combining EfficientNet-B3 with multi-layer perceptron (MLP) projection heads to preserve cross-modal alignment while significantly reducing computational requirements. We demonstrate that our distilled model achieves a balance between efficiency and performance, making it ideal for deployment in real-world scenarios. Experimental results show that Clip4Retrofit can perform real-time image labeling and object identification on edge devices with limited resources, offering a practical solution for applications such as autonomous driving and retrofitting existing systems. This work bridges the gap between state-of-the-art vision-language models and their deployment in resource-constrained environments, paving the way for broader adoption of foundation models in edge computing.
\end{abstract}    
\section{Introduction}
\label{sec:intro}

The rapid advancement of vision-language models, particularly OpenAI's CLIP (Contrastive Language--Image Pretraining) \cite{radford2021clip}, has enabled remarkable progress in zero-shot and few-shot learning tasks by aligning visual and textual representations in a shared embedding space. However, deploying such large-scale models on resource-constrained edge devices remains a significant challenge. Edge devices, such as the \textit{Retrofit} system used for in-car image and video collection, are often limited in computational power, memory, and energy efficiency, making it impractical to run models like CLIP in real-time \cite{han2022survey, tan2020efficientnet}. This limitation hinders the adoption of state-of-the-art vision-language models in practical applications, such as autonomous driving, where real-time image labeling and object identification are critical.

Real-time image labeling on edge devices has become increasingly important for applications like autonomous driving, where timely and accurate environment perception is essential for navigation and decision-making \cite{grigorescu2020survey}. The \textit{Retrofit} system, deployed on cars with limited memory and compute resource for image and video collection and processing , represents a prime use case for such technology. However, existing solutions often rely on lightweight models that sacrifice accuracy for efficiency or require cloud-based processing, which introduces latency and privacy concerns \cite{howard2020mobilenets, zhang2021edgeai}. By enabling efficient deployment of CLIP-like capabilities on edge devices, we can unlock new possibilities for real-time, on-device image labeling, reducing reliance on cloud infrastructure and improving system responsiveness.

In recent years, there have been tremendous advancements in efficient neural network architectures, driven by techniques such as pruning, quantization, and model distillation for vision tasks \cite{hinton2020distillation, zhu2022distillation}. However, applying these methods effectively to cross-modal architectures such as CLIP remains relatively underexplored. One critical challenge is maintaining cross-modal alignment—the essential semantic linkage between visual and textual embeddings—during the model distillation process \cite{jia2021scaling}.


To address these challenges, we propose \textit{Clip4Retrofit}, an efficient model distillation framework that transfers the knowledge of the CLIP model to a lightweight student model based on EfficientNet-B3 \cite{tan2021efficientnetv2}. Our approach incorporates multi-layer perceptron (MLP) projection heads to preserve cross-modal alignment while significantly reducing computational requirements. By systematically comparing various lightweight architectures, we demonstrate that EfficientNet-B3, combined with our distillation strategy, achieves an optimal balance between efficiency and performance. The distilled model is designed for deployment on the \textit{Retrofit} system, enabling real-time image labeling and object identification on edge devices with limited resources.

Experimental results show that \textit{Clip4Retrofit} achieves competitive performance in image labeling tasks while operating under strict computational constraints. On the \textit{Retrofit} system, our distilled model achieves real-time inference speeds with minimal accuracy degradation compared to the original CLIP model. These results highlight the potential of \textit{Clip4Retrofit} to bridge the gap between state-of-the-art vision-language models and their deployment in resource-constrained environments.

We summarize our main contributions as follows::
\begin{itemize}
    \item We introduce a distillation framework that preserves the cross-modal alignment capabilities of CLIP while significantly reducing model size and computational complexity.
    \item We propose a student model based on EfficientNet-B3 with MLP projection heads, optimized for edge device deployment.
    \item We demonstrate the practical applicability of our approach by deploying \textit{Clip4Retrofit} on the \textit{Retrofit} system, enabling real-time image labeling on in-car cameras.
    \item  We analyzed the trade-offs between accuracy, inference speed, and resource usage, showcasing the effectiveness of our approach in real-world scenarios.
\end{itemize}

\section{Related Work}
\subsection{Vision-Language Models}
Vision-language models, such as OpenAI's CLIP \cite{radford2021clip}, have transformed multimodal learning by enabling zero-shot and few-shot tasks through cross-modal alignment. Inspired by CLIP’s success, other models like ALIGN \cite{jia2021scaling}, BLIP \cite{li2022blip}, EVA-CLIP \cite{sun2023eva}, and Florence \cite{yuan2021florence} have leveraged large-scale pretraining on image-text pairs to achieve state-of-the-art performance. However, these models are computationally demanding and memory-intensive, making them impractical for deployment on edge devices. While recent efforts have explored lightweight alternatives \cite{tan2021efficientnetv2, papoudakis2025appvlm}, such models often compromise performance and remain unsuitable for hardware-constrained environments. To address this limitation, we propose a distillation-based approach to compress CLIP into a lightweight model that maintains its performance while being deployable on edge devices such as \textit{Retrofit}.

\subsection{Model Distillation}
Model distillation has emerged as a powerful technique for compressing large neural networks into smaller, more efficient models while preserving their performance. Recent advances in distillation have focused on transferring knowledge from large teacher models to lightweight student models, particularly in vision and language tasks \cite{hinton2020distillation, zhu2022distillation}. For vision-language models, distillation is particularly challenging due to the need to maintain cross-modal alignment between visual and textual embeddings. Although recent efforts by \cite{jia2021scaling} demonstrated the feasibility of distilling large vision-language models like CLIP, their approach remains computationally expensive for edge deployment. Our work builds on these foundations by introducing a novel distillation strategy tailored for resource-constrained devices like the \textit{Retrofit} system.

\subsection{Edge Device Deployment}
Deploying deep learning models on resource-constrained edge devices has been a major focus of research in recent years. Techniques such as model quantization \cite{zhang2021edgeai}, pruning \cite{han2022survey}, and knowledge distillation \cite{hinton2020distillation} have been widely adopted to reduce model size and computational requirements. However, most existing methods focus on single-modality models, such as convolutional neural networks (CNNs) for image classification \cite{howard2020mobilenets}. Deploying multimodal models like CLIP on edge devices remains underexplored, particularly in real-time applications such as autonomous driving. Our work bridges this gap by enabling efficient deployment of a distilled CLIP model on the \textit{Retrofit} system, a real-world edge device for in-car image and video collection.

\subsection{Automatic Image Labeling}
Automatic image labeling is a critical task for applications such as autonomous driving, where real-time environment perception is essential. Traditional approaches rely on supervised learning with large annotated datasets \cite{grigorescu2020survey}, but these methods are often computationally expensive and require significant human effort. Recent advances in self-supervised and weakly supervised learning have reduced the need for labeled data \cite{jia2021scaling}, but these methods still struggle with real-time performance on edge devices. Our work addresses this challenge by leveraging the zero-shot capabilities of CLIP, distilled into a lightweight model that can perform real-time image labeling on the \textit{Retrofit} system.

\subsection{Gaps and Limitations}
Despite significant progress in model distillation, vision-language models, and edge device deployment, several gaps remain. First, existing distillation methods for vision-language models often fail to preserve cross-modal alignment, limiting their applicability to tasks like image labeling. Second, while lightweight models like EfficientNet \cite{tan2021efficientnetv2} have been widely adopted for edge deployment, they are not designed for multimodal tasks. Finally, there is a lack of research on deploying vision-language models in real-world edge computing scenarios, such as autonomous driving. Our work addresses these gaps by introducing a novel distillation framework that preserves cross-modal alignment, leverages lightweight architectures, and enables real-time deployment on the \textit{Retrofit} system.
\section{Methodology}
Our framework distills knowledge from the CLIP teacher model into a lightweight student model designed for deployment on the \textit{Retrofit} edge device. The student model combines an EfficientNet-B3 backbone with multi-layer perceptron (MLP) projection heads to align with CLIP's cross-modal embedding space. During training, we minimize the \textit{CosineEmbeddingLoss} between the student's and teacher's output embeddings (1x768 vectors), while leveraging mixed-precision training and gradient checkpointing to reduce memory usage. After training, the model is converted to ONNX format, optimized via quantization and pruning, and compiled into a binary model for deployment on the \textit{Retrofit} camera. The final model size is approximately 25 MB, a significant reduction from the original CLIP-L/14 model size of 1.2 GB, enabling real-time inference on resource-constrained hardware.
\subsection{Preliminary}

\textbf{CLIP Model} CLIP (Contrastive Language–Image Pretraining) \cite{radford2021clip} is a vision-language model trained on 400 million image-text pairs to align visual and textual representations in a shared embedding space. It consists of dual encoders: a Vision Transformer (ViT) \cite{dosovitskiy2020image} or ResNet-based \cite{he2016deep} image encoder and a transformer-based text encoder. CLIP enables zero-shot image classification by computing the cosine similarity between image and text embeddings. Despite its versatility, the model's computational demands (e.g., ViT-L/14 requires 304M parameters) make it unsuitable for edge deployment.

\textbf{EfficientNet Model} EfficientNet \cite{tan2020efficientnet} is a family of convolutional neural networks optimized for accuracy and efficiency through compound scaling of depth, width, and resolution. EfficientNet-B3, the backbone of our student model, achieves a balance between performance and computational cost with 12M parameters and 1.8B FLOPs for $300 \times 300$ inputs. Its scalability and hardware efficiency have made it a popular choice for edge applications \cite{tan2021efficientnetv2}.

\textbf{Lightweight Vision Transformers} Recent work has adapted Vision Transformers (ViTs) for edge devices by reducing their computational complexity. Light-ViT \cite{liu2022lightvit} replaces self-attention with dynamic position-aware convolutions, achieving 80\% top-1 accuracy on ImageNet with only 1.3G FLOPs. Similarly, MobileViT \cite{mehta2022mobilevit} integrates CNNs and ViTs for mobile-friendly inference. While these models offer promising efficiency, their performance in cross-modal tasks remains inferior to CLIP-based architectures in our experiments.

\subsection{Distillation Framework}

\begin{figure*}[t]
  \includegraphics[width=\textwidth]{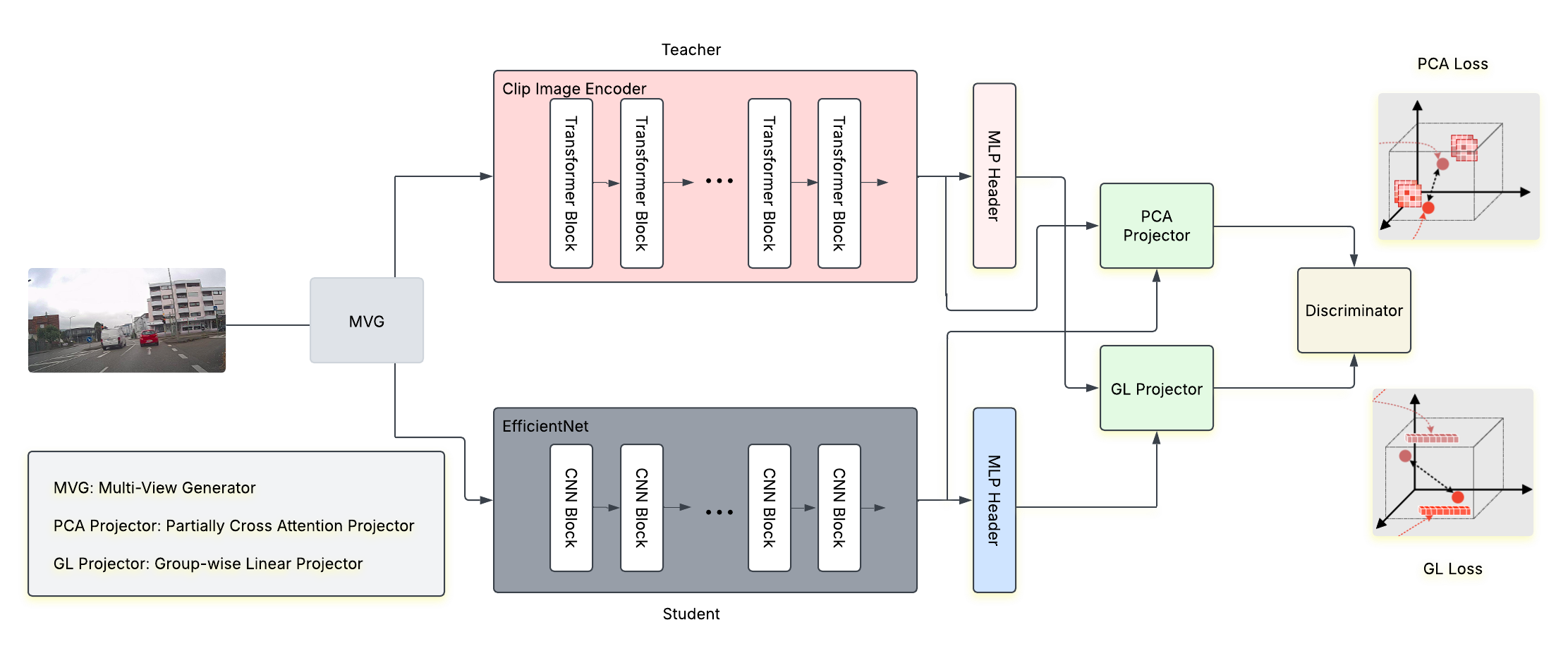} 
  \caption{Overall framework of the knowledge distillation method}
  \label{fig:framework}
\end{figure*}

\subsection{Model Architecture}
Our knowledge distillation framework is designed to transfer knowledge from a pre-trained CLIP Vision Transformer (ViT) teacher model to a lightweight EfficientNet-based student model. The framework consists of two main components: the teacher model (CLIP) and the student model (EfficientNet). The student model is based on EfficientNet-B3 \cite{tan2020efficientnet}, augmented with a 3-layer MLP projection head to match CLIP's embedding dimension (768). The teacher model is a pre-trained CLIP ViT, which provides high-quality embeddings for guiding the student. Figure \ref{fig:framework} illustrates the overall architecture.

\subsubsection{Feature Alignment via Cross-Architecture Projectors}
The key challenge in distilling knowledge from a Transformer-based teacher to a convolutional student lies in the architectural mismatch between the two models. To address this, we extend the cross architecture distillation method introduced in \cite{liu2022cross}, where knowledge transfer is optimized despite architectural disparities between teacher and student through two feature alignment mechanisms: the \textbf{Partially Cross Attention (PCA) Projector} and the \textbf{Group-wise Linear (GL) Projector}.

\begin{itemize}
    \item \textbf{Partially Cross Attention (PCA) Projector}: 
    This module aligns the student's spatial features with the teacher's self-attention mechanism. The student's feature map $\mathbf{h}_S \in \mathbb{R}^{C \times H'W'}$ is transformed into query ($Q_S$), key ($K_S$), and value ($V_S$) matrices using a series of $3 \times 3$ convolutional layers:
    \begin{equation}
        \{Q_S, K_S, V_S\} = \text{Conv}(\mathbf{h}_S).
    \end{equation}
    The student's attention matrix is computed as:
    \begin{equation}
        \text{Attn}_S = \text{softmax} \left( \frac{Q_S K_S^T}{\sqrt{d}} \right) V_S,
    \end{equation}
    where $d$ is the dimension of the query. The PCA loss minimizes the discrepancy between the student's and teacher's attention maps:
    \begin{equation}
        \mathcal{L}_{\text{PCA}} = \|\text{Attn}_T - \text{Attn}_S\|_2^2.
    \end{equation}
    This alignment encourages the student to learn global relational features from the teacher, despite its convolutional architecture.

    \item \textbf{Group-wise Linear (GL) Projector}: 
    This module bridges the gap between the student's spatial feature maps and the teacher's token embeddings. The student's features are projected into the teacher's embedding space using a fully connected layer:
    \begin{equation}
        \mathbf{h}_S' = \text{FC}(\mathbf{h}_S),
    \end{equation}
    where $\text{FC}(\cdot)$ represents a group-wise linear transformation. Instead of using an L2 loss, we employ \textbf{cosine embedding similarity} to measure the alignment between the teacher's and student's embeddings. The cosine embedding loss is defined as:
    \begin{equation}
        \mathcal{L}_{\text{GL}} = 1 - \frac{\mathbf{h}_T \cdot \mathbf{h}_S'}{\|\mathbf{h}_T\| \|\mathbf{h}_S'\|},
    \end{equation}
    where $\mathbf{h}_T$ and $\mathbf{h}_S'$ are the teacher's and student's embeddings, respectively. This loss ensures that the student's embeddings are directionally aligned with the teacher's, enabling effective knowledge transfer across architectures.
\end{itemize}

\subsubsection{Multi-View Robust Training}
To enhance the robustness and generalization of the student model, we employ a \textbf{multi-view robust training scheme}. Given an input image $\mathbf{x}$, we generate augmented views $\tilde{\mathbf{x}} = g(\mathbf{x})$ using random transformations such as cropping, masking, and color jittering. These transformations simulate diverse real-world conditions, ensuring that the student model remains robust to input variations.

The student produces perturbed feature representations $\tilde{\mathbf{h}}_S$, and a discriminator $D$ is trained to distinguish between teacher embeddings $\mathbf{h}_T$ and student embeddings $\tilde{\mathbf{h}}_S$. The adversarial training loss is defined as:
\begin{equation}
    \mathcal{L}_{\text{adv}} = \mathbb{E}[-\log D(\mathbf{h}_T) - \log(1 - D(\tilde{\mathbf{h}}_S))].
\end{equation}
Minimizing this loss encourages the student to produce features that are indistinguishable from the teacher's, improving its ability to generalize to unseen data. The discriminator is implemented as a lightweight neural network with three fully connected layers, ensuring minimal computational overhead.

\subsubsection{Optimization}
The overall distillation objective combines the PCA, GL, and adversarial losses:
\begin{equation}
    \mathcal{L}_{\text{total}} = \mathcal{L}_{\text{PCA}} + \mathcal{L}_{\text{GL}} + \lambda \mathcal{L}_{\text{adv}}
    \end{equation}
where $\lambda$ is a hyperparameter that balances the contribution of the adversarial loss. The model is optimized in an alternating fashion: the student minimizes $\mathcal{L}_{\text{total}}$, while the discriminator maximizes $\mathcal{L}_{\text{adv}}$. This adversarial training process ensures that the student learns robust and generalizable representations.

After training, the PCA and GL projectors are removed, leaving a lightweight and efficient student model suitable for deployment on edge devices. The final model retains the ability to perform real-time image labeling while maintaining high accuracy and computational efficiency.

We evaluated alternative student architectures, including EfficientNet-B4, EfficientNet-B5, and lightweight Vision Transformers (Light-ViT) \cite{liu2022lightvit}, but EfficientNet-B3 achieved the best trade-off between accuracy and inference speed.

To optimize memory efficiency, we employ mixed-precision training (FP16/FP32) and gradient checkpointing. Additionally, only the final six layers of the student model are fine-tuned during training, while all preceding layers remain frozen. This approach reduces computational overhead while retaining the pretrained feature extraction capabilities of EfficientNet-B3.

\subsection{Deployment Pipeline}
\begin{figure}[htbp]
    \centering
  \includegraphics[width=0.5\textwidth]{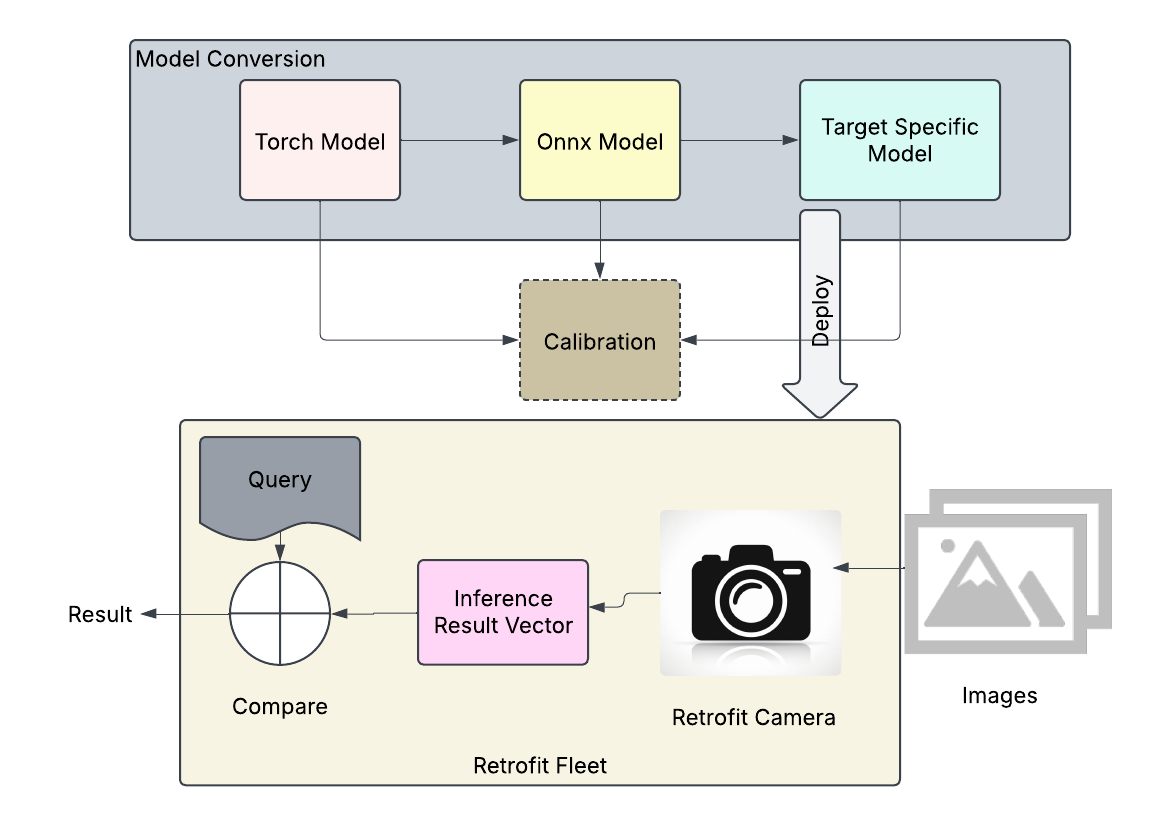} 
  \caption{Pipeline of the deployment of distilled model on the Retrofit camera}
  \label{fig:deploy}
\end{figure}
After training, the PyTorch model is exported to ONNX format to enable hardware-agnostic deployment. This conversion ensures compatibility with a wide range of hardware platforms, including the Retrofit camera system. To mitigate performance degradation during conversion, we perform calibration using a held-out validation set, ensuring numerical precision and alignment between the original and converted models. The ONNX model is further optimized by replacing computationally heavy operators (e.g., matrix multiplications) with hardware-friendly equivalents and quantizing weights to INT16 precision.

The optimized ONNX model is then compiled into a binary model file using vendor-specific tools provided with the \textit{Retrofit} camera SDK. The binary model is designed for efficient execution on the Retrofit hardware for accelerated inference. 

\subsubsection{Inference on the Retrofit Camera}
The Retrofit camera continuously captures images and videos from its environment. For each incoming frame, the binary model performs inference, generating embedding vectors that represent the semantic content of the input images. These vectors are compared with a set of pre-stored query embeddings, which correspond to specific objects or scenes of interest. The comparison is performed using cosine similarity, ensuring robust and efficient matching even in diverse real-world conditions.

Based on the similarity scores, the system generates inference results, such as object identification or scene classification. These results are used to support real-time decision-making in autonomous driving scenarios, such as detecting traffic lights, recognizing pedestrians, or identifying road signs. The entire pipeline operates with an inference runtime of approximately \textbf{35 milliseconds (ms)} per frame, meeting the real-time requirements for edge deployment.

\subsubsection{Deployment on the Retrofit Fleet}
The optimized binary model is deployed across the Retrofit fleet, enabling scalable and efficient real-time image labeling on edge devices. Each camera in the fleet operates independently, performing local inference and generating results without relying on cloud-based processing. This decentralized approach ensures low latency, reduces bandwidth requirements, and enhances privacy by keeping sensitive data on the device.

\subsubsection{Calibration and Validation}
To ensure the accuracy and reliability of the deployed model, we perform extensive calibration and validation. The calibration process involves fine-tuning the model's quantization parameters using a held-out validation set, ensuring minimal performance degradation during conversion. Additionally, we validate the model on real-world driving scenarios, including adverse weather conditions and low-light environments, to ensure robustness and generalization.

The deployment pipeline, from model conversion to real-time inference, is designed to maximize efficiency and accuracy while minimizing resource usage. This makes our distilled CLIP model a practical and scalable solution for real-time image labeling on edge devices in autonomous driving applications.
\subsubsection{Hardware}
The Retrofit camera is a high-performance edge device designed for automotive applications. It features a front-facing camera capable of capturing 1080p video at up to 30 frames per second (fps) with a 90° horizontal field of view (HFOV). The camera is powered by a quad-core CPU and a Neural Processing Unit (NPU) delivering 3 eTOPS of compute power for efficient deep learning inference. Additionally, the system is equipped with a 4G modem, GPS, and a 6-axis IMU, enabling robust perception and localization capabilities for autonomous driving scenarios. This hardware platform provides the computational resources necessary for real-time deployment of our distilled CLIP model, ensuring low-latency and energy-efficient operation.
\subsubsection{Edge Device Performance}
For input images of size $300 \times 300$, the optimized model achieves an inference runtime of approximately \textbf{35 milliseconds (ms)} per frame.  The binary model itself occupies \textbf{24.6 MB} of storage, making it lightweight and suitable for deployment on resource-constrained devices. And the \textbf{binary model memory requirement} for each perception task is \textbf{25.6 MB}, ensuring efficient utilization of limited hardware resources.

\section{Experiment}
\subsection{Experiment Settings}
\subsubsection{Dataset}  
\begin{itemize}
    \item Internal Dataset: We constructed a real-world driving dataset using images captured by BOSCH-equipped vehicles between 2022 and 2023. To optimize this dataset for efficient model distillation, we implemented a robust automated data curation pipeline, drawing inspiration from contemporary vision-language model training strategies \cite{dino2023}. Initially, we employed cosine similarity-based clustering \cite{pizzi2022self} to filter and deduplicate images, effectively eliminating near-duplicates while preserving scene diversity. Subsequently, a self-supervised retrieval system was leveraged to augment the dataset. This involved retrieving the top-4 nearest visual neighbors from a larger, uncurated image pool and clustering them into 1000 groups using k-means \cite{johnson2019billion}. This strategic curation process yielded a high-quality dataset tailored for real-time model distillation, enabling the student model to learn robust and generalizable representations from data.
    \item Public Dataset: In addition to the internal Bosch Cars dataset, we evaluate our framework on two widely used public benchmarks: the Cityscapes dataset \cite{cordts2016cityscapes} and the Mapillary Vistas dataset v2.0 \cite{neuhold2017mapillary}.  Cityscapes includes high-resolution images ($2048 \times 1024$) from diverse urban environments, accompanied by detailed pixel-level annotations for 34 semantic classes, such as vehicles, pedestrians, and road infrastructure. The Mapillary Vistas Dataset v2.0 further enriches our evaluation by providing globally sourced, high-resolution images annotated across 124 semantic categories, encompassing a broader spectrum of urban and suburban conditions. For both datasets, semantic mask annotations within each image are treated as labels for the entire image. We conduct zero-shot evaluations on the provided validation sets. This comprehensive evaluation ensures our framework's generalization capability and robustness across diverse real-world scenarios, beyond the specific context of the Bosch Cars dataset.
\end{itemize}

\subsubsection{Metrics}
We used the Receiver Operating Characteristic (ROC) curve and Area Under the Curve (AUC) as primary metrics. The ROC curve visualizes the trade-off between True Positive Rate (TPR) and False Positive Rate (FPR) across different confidence thresholds, while the AUC summarizes this curve into a single value representing the model's discriminative ability. We also established reference AUC values for each class based on the original CLIP models to contextualize the performance of our distilled model. These metrics provide a comprehensive evaluation of the model's ability to handle real-world driving scenarios, such as identifying trucks, tunnels, bicycles, bridges, and traffic lights.

\subsection{Results}
\begin{figure}[htbp]
    \centering
    \begin{subfigure}[b]{0.23\textwidth}  
        \includegraphics[width=\textwidth]{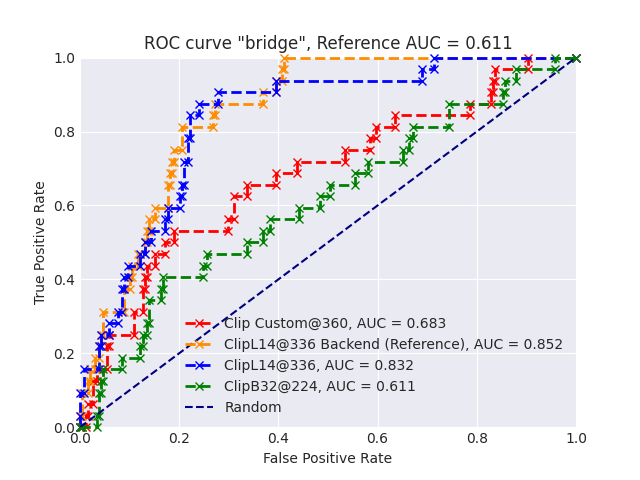}
        \caption{ROC curve for "bridge" class.}
        \label{fig:bridge}
    \end{subfigure}
    \hfill
    \begin{subfigure}[b]{0.23\textwidth}
        \includegraphics[width=\textwidth]{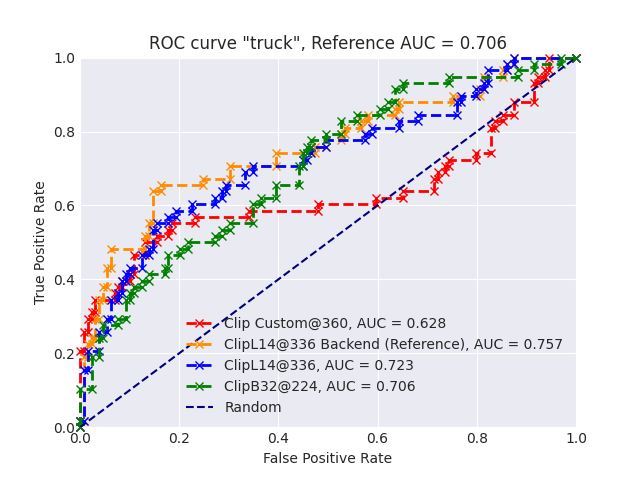}
        \caption{ROC curve for "truck" class.}
        \label{fig:truck}
    \end{subfigure}
    \hfill
    \begin{subfigure}[b]{0.21\textwidth}
        \includegraphics[width=\textwidth]{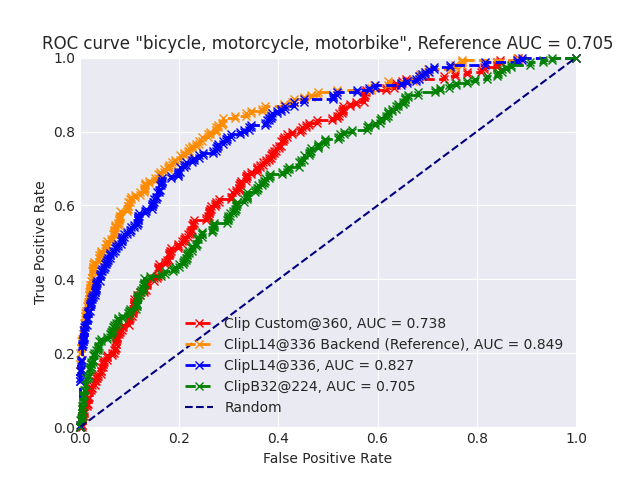}
        \caption{ROC curve for "bicycle, motorcycle, motorbike" class.}
        \label{fig:traffic_lights}
    \end{subfigure}
    \hfill
    \begin{subfigure}[b]{0.23\textwidth}
        \includegraphics[width=\textwidth]{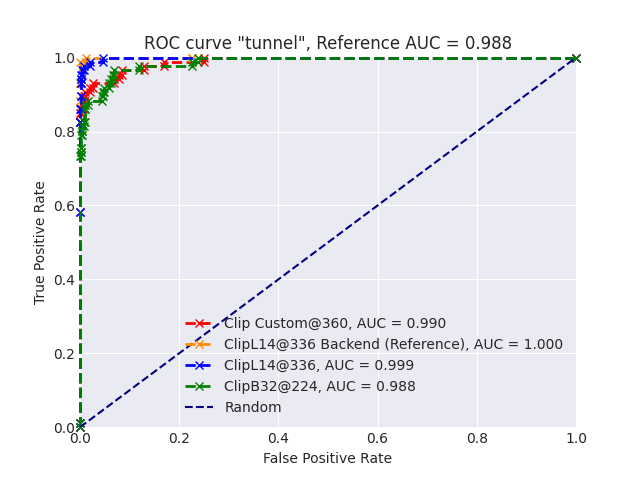}
        \caption{ROC curve for "tunnel" class.}
        \label{fig:tunnel}
    \end{subfigure}
    \caption{ROC curves comparing the performance of the distilled CLIP model (Clip Custom) with the original CLIP models (ClipL14 and ClipB32) on the Bosch Cars dataset.}
    \label{fig:roc_curves}
\end{figure}

\begin{figure}[htbp]
    \centering
    \begin{subfigure}[b]{0.21\textwidth}  
        \includegraphics[width=\textwidth]{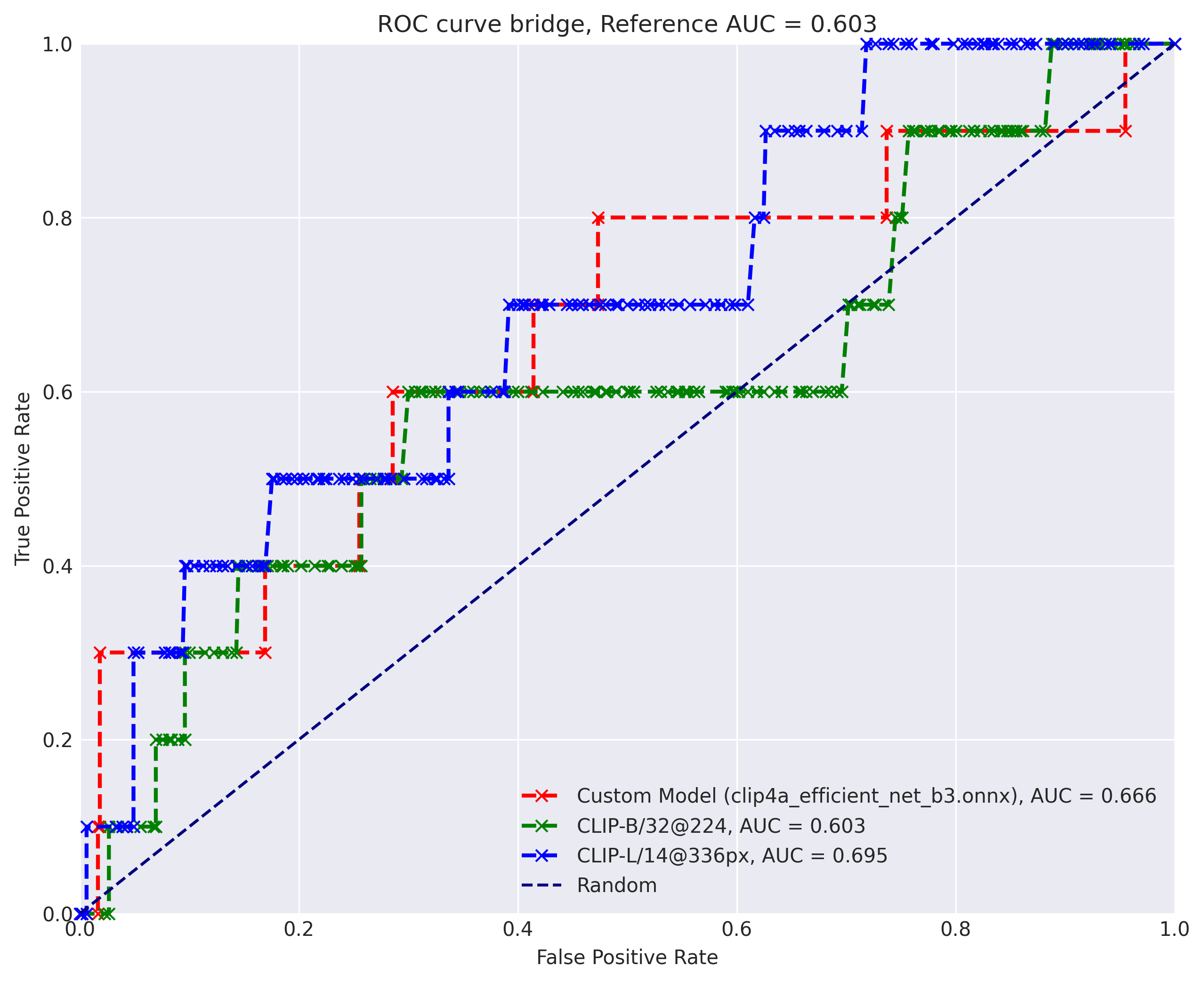}
        \caption{ROC curve for "bridge" class.}
        \label{fig:bridge_cityscapes}
    \end{subfigure}
    \hfill
    \begin{subfigure}[b]{0.21\textwidth}
        \includegraphics[width=\textwidth]{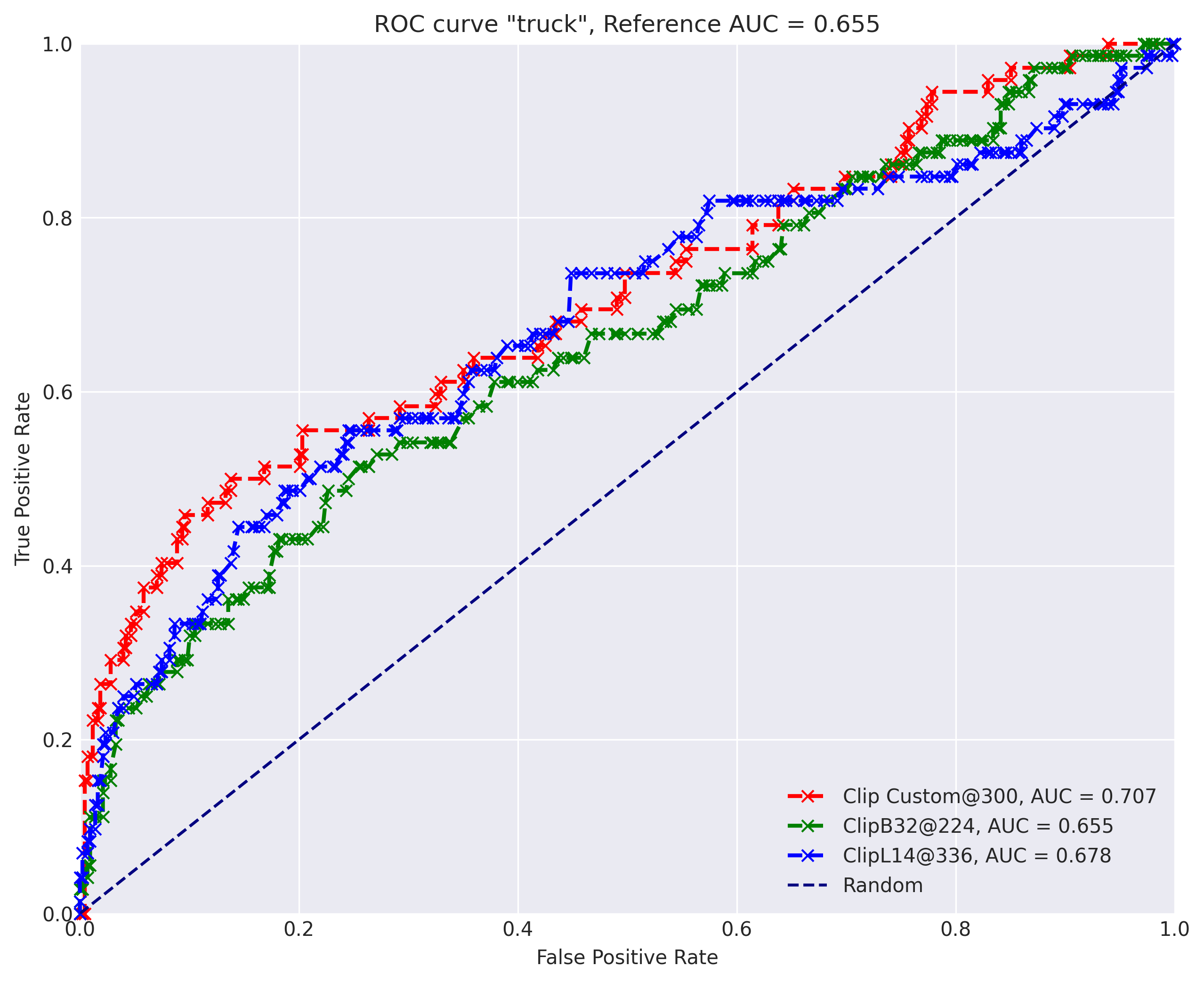}
        \caption{ROC curve for "truck" class.}
        \label{fig:truck_cityscapes}
    \end{subfigure}
    \hfill
    \begin{subfigure}[b]{0.2\textwidth}
        \includegraphics[width=\textwidth]{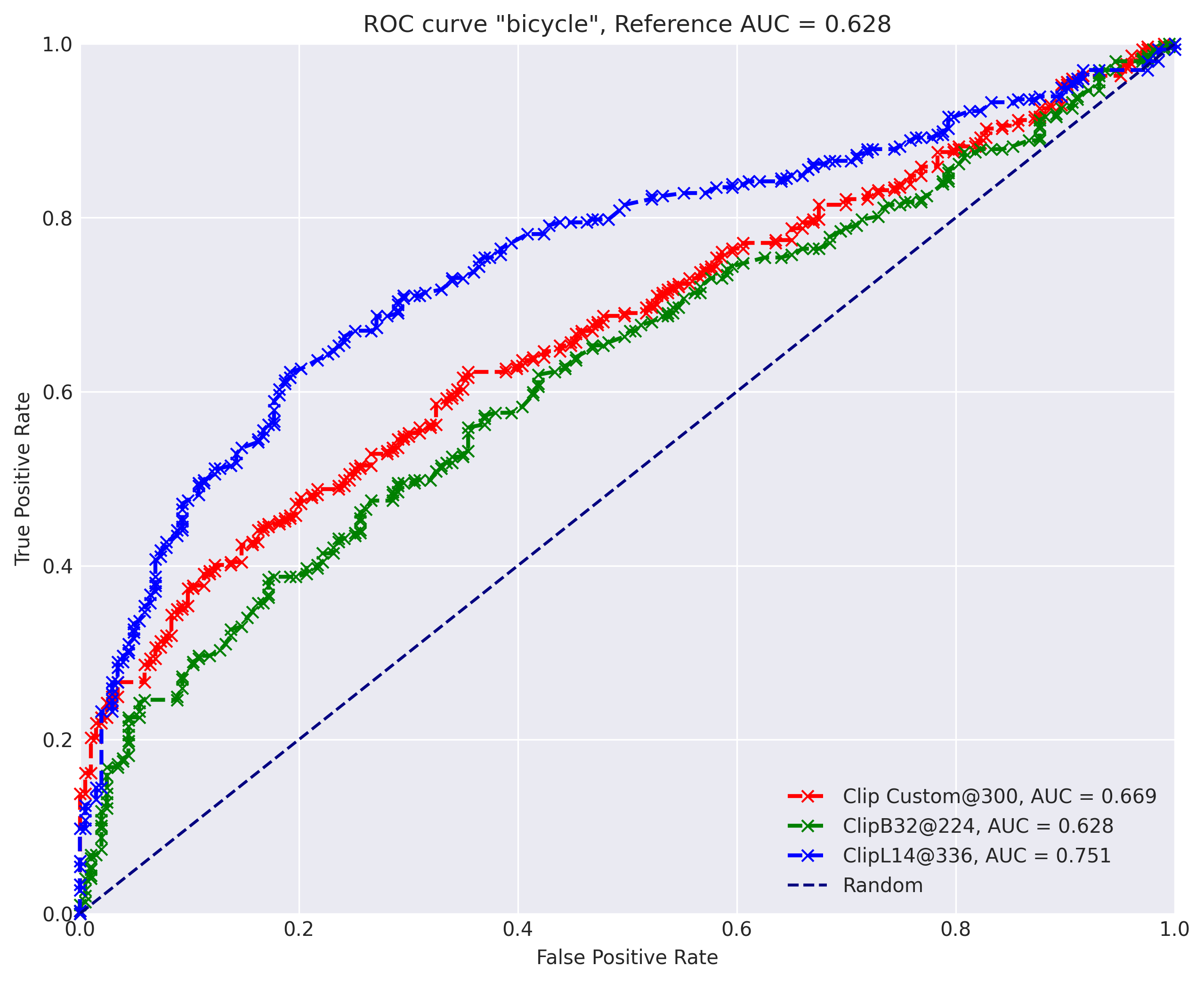}
        \caption{ROC curve for "bicycle" class.}
        \label{fig:bike_cityscapes}
    \end{subfigure}
    \hfill
    \begin{subfigure}[b]{0.21\textwidth}
        \includegraphics[width=\textwidth]{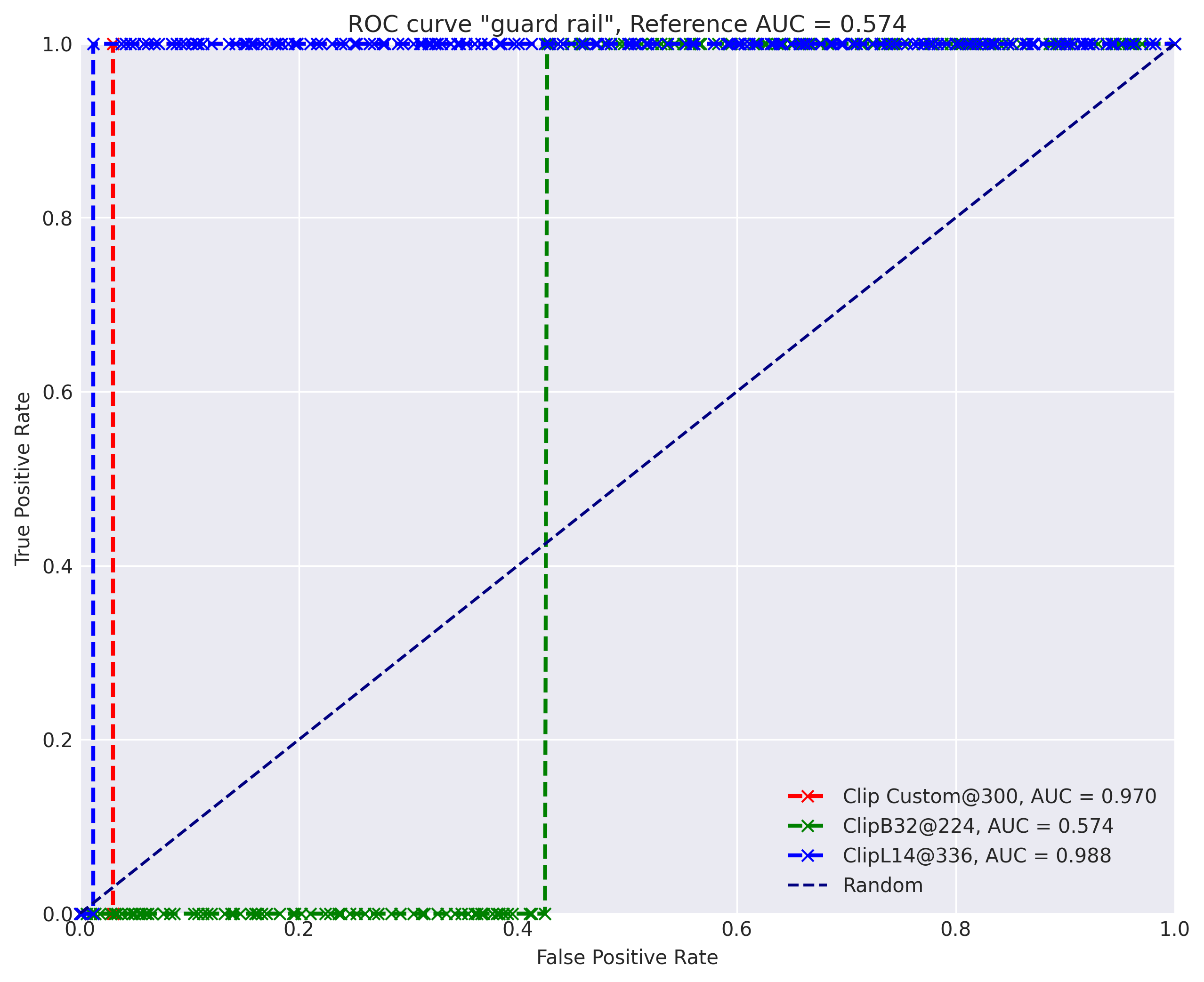}
        \caption{ROC curve for "guard rail" class.}
        \label{fig:guard_rail_cityscapes}
    \end{subfigure}
    \caption{ROC curves comparing the performance of the distilled CLIP model (Clip Custom) with the original CLIP models (ClipL14 and ClipB32) on the Cityscapes dataset.}
    \label{fig:roc_curves_cityscapes}
\end{figure}

\begin{figure}[htbp]
    \centering
    \begin{subfigure}[b]{0.21\textwidth}  
        \includegraphics[width=\textwidth]{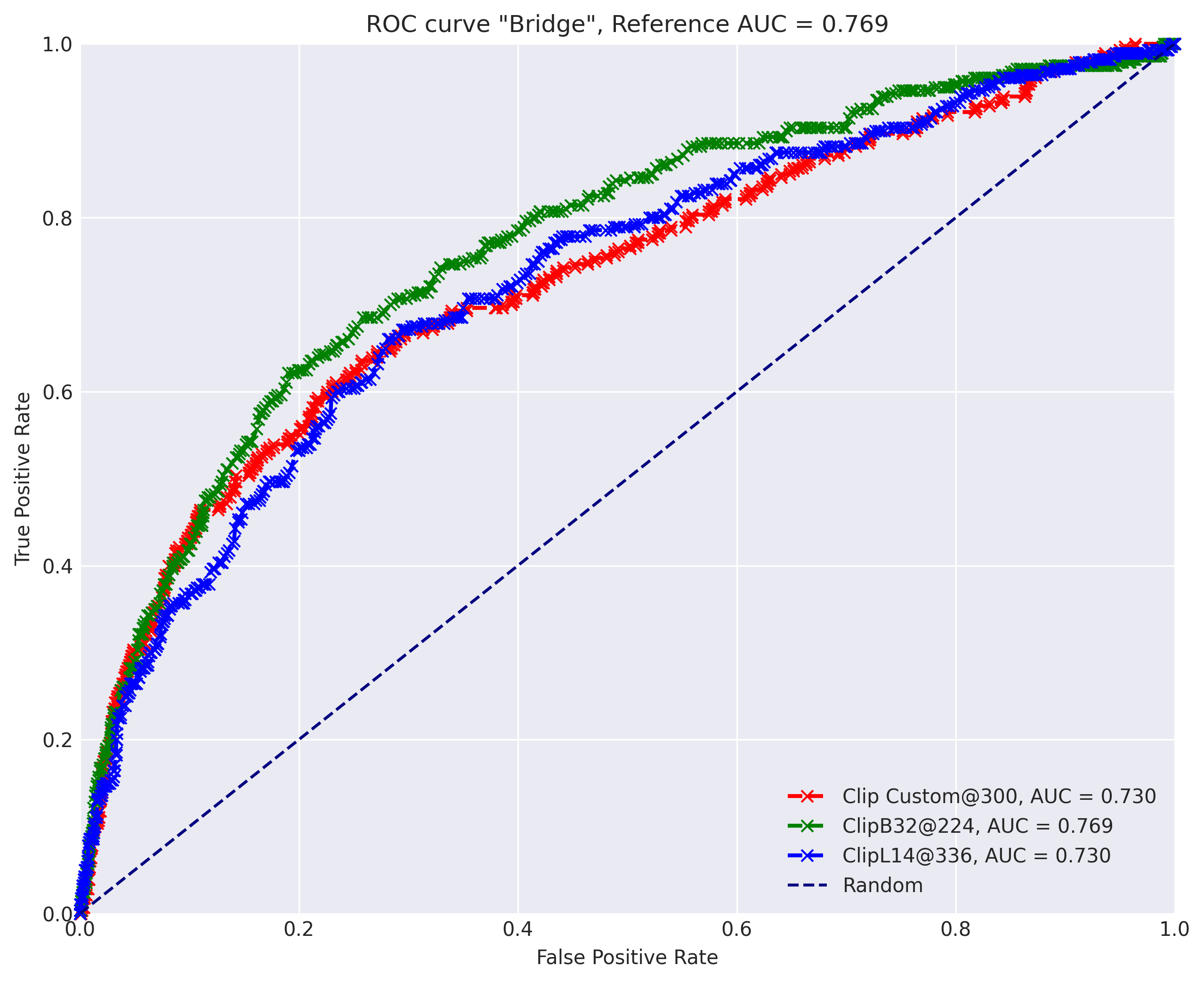}
        \caption{ROC curve for "bridge" class.}
        \label{fig:bridge_mapillary}
    \end{subfigure}
    \hfill
    \begin{subfigure}[b]{0.21\textwidth}
        \includegraphics[width=\textwidth]{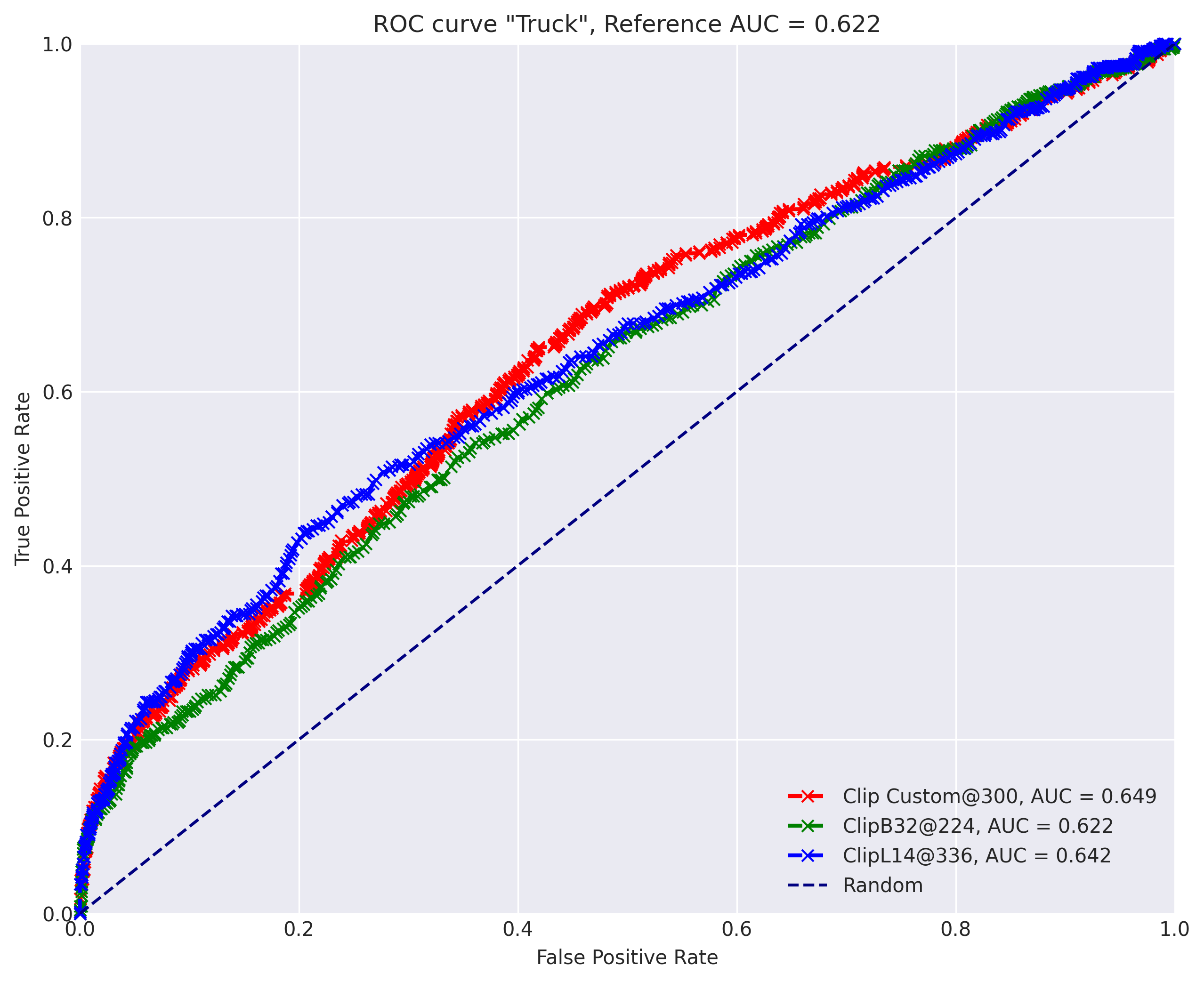}
        \caption{ROC curve for "truck" class.}
        \label{fig:truck_mapillary}
    \end{subfigure}
    \hfill
    \begin{subfigure}[b]{0.2\textwidth}
        \includegraphics[width=\textwidth]{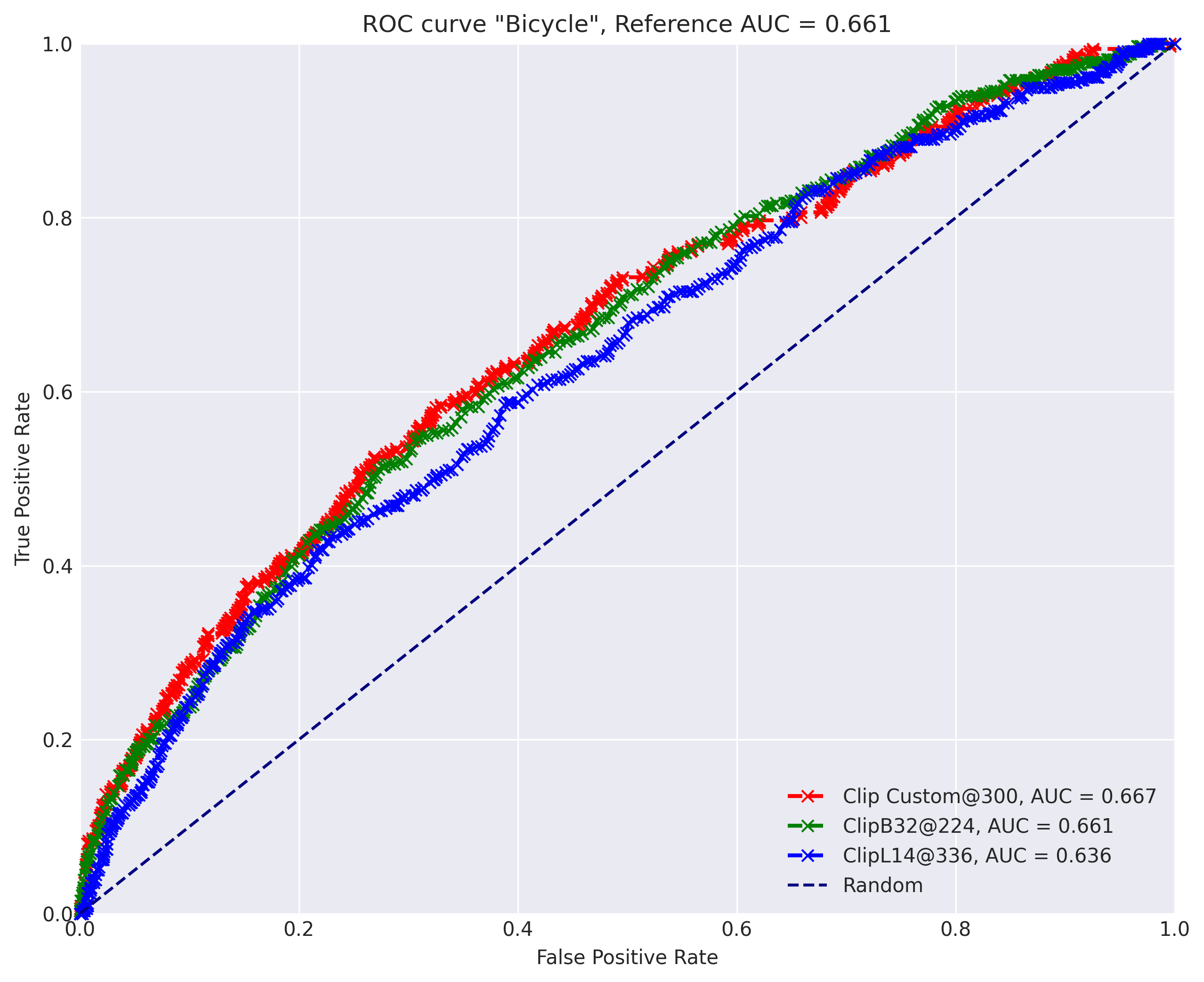}
        \caption{ROC curve for "bicycle" class.}
        \label{fig:bike_mapillary}
    \end{subfigure}
    \hfill
    \begin{subfigure}[b]{0.21\textwidth}
        \includegraphics[width=\textwidth]{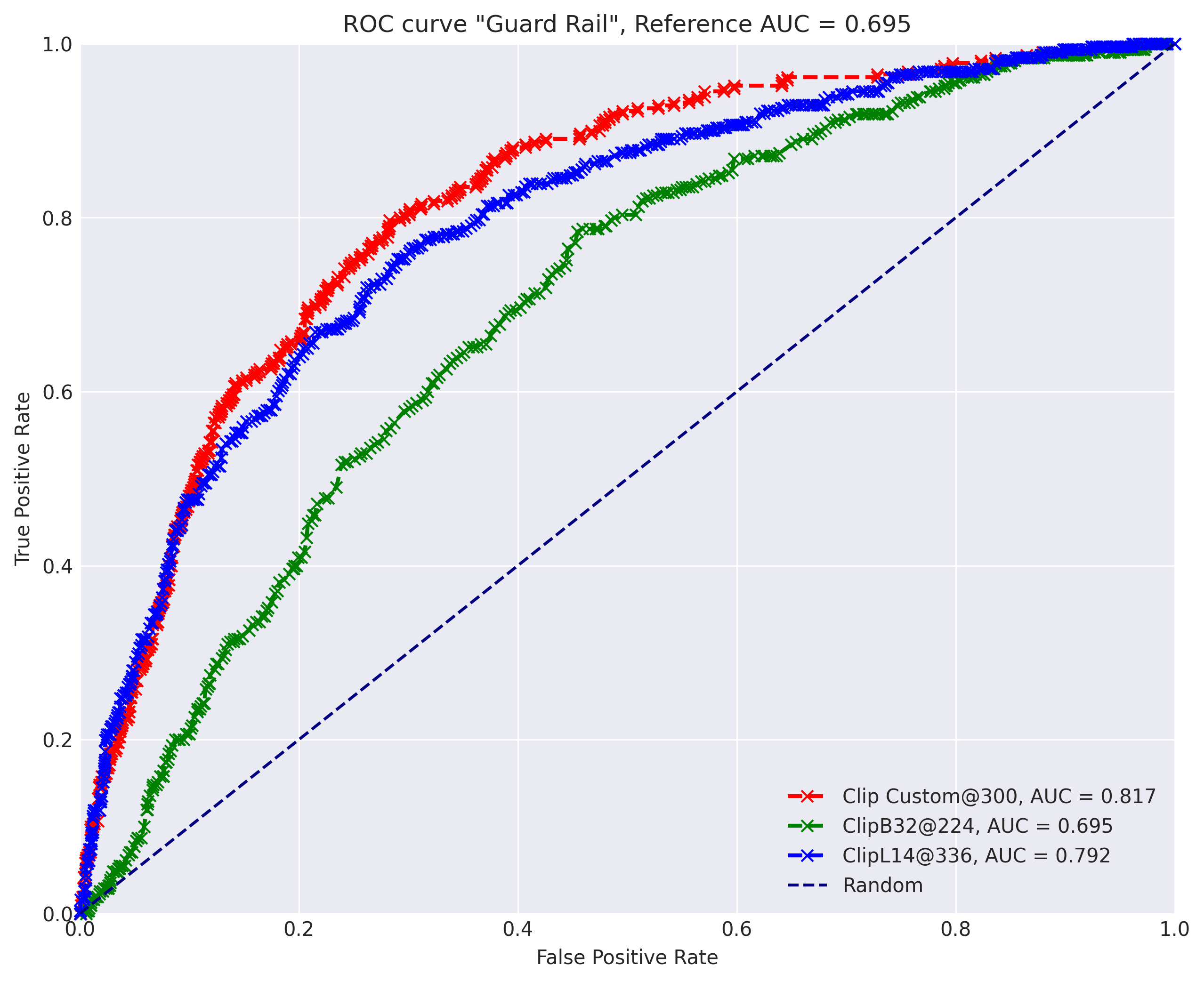}
        \caption{ROC curve for "guard rail" class.}
        \label{fig:guard_rail_mapillary}
    \end{subfigure}
    \caption{ROC curves comparing the performance of the distilled CLIP model (Clip Custom@300) with the original CLIP models (ClipL14@336 and ClipB32@224) on the Mapillary Vistas v2.0 dataset.}
    \label{fig:roc_curves_mapillary}
\end{figure}

Our extensive evaluation demonstrates the effectiveness and robustness of Clip4Retrofit across various challenging real-world datasets while being real-time. Figure \ref{fig:roc_curves}, Figure \ref{fig:roc_curves_cityscapes}), Figure \ref{fig:roc_curves_mapillary} illustrate the ROC curves comparing the Clip4Retrofit model against original CLIP models (ClipL14@336 and ClipB32@224) on Bosch Cars Dataset, CityScapes \cite{cordts2016cityscapes}, and Mapillary Vistas v2 \cite{neuhold2017mapillary} respectively. 

On Bosch Cars Dataset, Clip4Retrofit competitively labels images that contain objects and scene such as "bridge," "truck," "bike," and "tunnel," closely matching several original CLIP models.  Similarly, on the Cityscapes and Mapillary, Clip4Retrofit achieves even better performance in classes such as truck than baseline CLIP models (ViT-B/32 and ViT-L/14), while maintaining competitive performances in other classes.
Table \ref{table:evaluation_auc} summarizes the quantitative performance across multiple classes, models, and datasets, showing that Clip4Retrofit remains highly competitive with original CLIP models.
\begin{table*}[htbp]
\centering
\caption{Zero-Shot ROC-AUC Evaluation on Cityscapes and Mapillary Vistas}
\label{table:evaluation_auc}
\resizebox{\linewidth}{!}{
\begin{tabular}{l|ccc|ccc}
\toprule
Semantic Class & \multicolumn{3}{c|}{Cityscapes} & \multicolumn{3}{c}{Mapillary Vistas v2.0} \\ 
               & Clip4Retrofit & ViT-B/32 & ViT-L/14 & Clip4Retrofit & ViT-B/32 & ViT-L/14 \\ 
\midrule
Bicycle      & 0.6692 & 0.6277 & \textbf{0.7512} & \textbf{0.6672} & 0.6609 & 0.6356 \\
Bridge      & 0.6657 & 0.6031 & \textbf{0.6947} & 0.7299 & \textbf{0.7695} & 0.7298 \\
Building     & 0.7199 & \textbf{0.7531} & 0.7233 & 0.7428 & \textbf{0.7997} & 0.7131 \\
Bus          & \textbf{0.7765} & 0.6550 & 0.6777 & 0.6527 & \textbf{0.6610} & 0.6487 \\
Fence        & \textbf{0.6756} & 0.5210 & \textbf{0.6756} & 0.5452 & \textbf{0.5786} & 0.5568 \\
Guard Rail   & 0.9699 & 0.5741 & \textbf{0.9880} & \textbf{0.8174} & 0.6946 & 0.7924 \\
Truck        & \textbf{0.7066} & 0.6550 & 0.6777 & \textbf{0.6490} & 0.6223 & 0.6422 \\
Vegetation	& \textbf{0.6900} &	0.5101 &	0.6554	&0.5973	& \textbf{0.6812}	&0.5726 \\
\bottomrule
\end{tabular}
}
\end{table*}
Overall,  the analysis confirms that Clip4Retrofit provides consistent and robust zero-shot visual recognition across multiple urban scenarios, while also being significantly faster, validating our approach as a practical solution for real-world onboard deployment in automotive and urban scene understanding applications.

\section{Conclusion and Future Work}
In this work, we introduced Clip4Retrofit, a novel model distillation framework that enables real-time image labeling and object recognition on resource-constrained edge devices. By distilling knowledge from CLIP into a lightweight EfficientNet-B3 with MLP projection heads, we significantly reduce computational overhead while preserving the model’s zero-shot recognition capabilities. Our extensive evaluations demonstrate that Clip4Retrofit achieves competitive accuracy while being efficient enough for deployment on Retrofit, an edge device designed for autonomous data collection in real-world scenarios. These results highlight the feasibility of leveraging vision-language models in edge AI applications, providing an efficient and scalable approach for automated data annotation.

Despite its advantages, our approach has some limitations. The distilled model, while lightweight, still exhibits performance degradation compared to the original CLIP model, particularly for complex multi-modal reasoning tasks. Additionally, the distillation process requires careful tuning of the teacher-student training pipeline, which may not generalize optimally across different datasets.

Future work will focus on extending our framework in several directions. First, we aim to explore more powerful vision-language models for distillation, particularly those tailored to automotive use cases. While the teacher model CLIP is generic and excels in broad vision-language tasks, it lacks domain-specific knowledge for automotive scenarios. For example, CLIP recognizes common objects like zebra crossings but struggles with less frequent road signs (e.g., specific traffic signs) or specialized traffic situations (e.g., zipper merging). Adapting or fine-tuning CLIP for automotive-specific tasks could significantly enhance the zero-shot capabilities of Clip4Retrofit. Second, we will investigate adaptive distillation techniques, allowing for dynamic model optimization based on available edge resources. This would enable Clip4Retrofit to adjust its computational requirements in real time, improving efficiency and scalability across diverse edge devices. 
Third, we plan to address the limitation of input resolution. Currently, the model processes images at a resolution of $300\times 300$, which achieves strong performance for large objects (e.g., bridges, trucks) but struggles with small or distant objects (e.g., lost cargo, open car doors at a distance). Exploring higher-resolution inputs or multi-scale feature extraction techniques could improve performance for these challenging cases. 
{
    \small
    \bibliographystyle{ieeenat_fullname}
    \bibliography{main}
}


\end{document}